\documentclass[journal]{IEEEtran}

\usepackage[utf8]{inputenc}         
\usepackage[T1]{fontenc}            
\usepackage{hyperref}               
\usepackage{url}                    
\usepackage{nicefrac}               
\usepackage{microtype}              
\usepackage{amsfonts}               
\usepackage{amsmath}                
\usepackage[dvipsnames]{xcolor}     
\usepackage{float}                  
\usepackage{graphicx}               
\graphicspath{{images/}}
\DeclareGraphicsExtensions{.pdf,.PDF,.jpg,.JPG,.jpeg,.JPEG,.png,.PNG}
\usepackage[caption=false]{subfig}  
\usepackage{booktabs}               
\usepackage{multicol}               
\usepackage{multirow}               
\usepackage{booktabs}               
\usepackage{array}                  
\newcolumntype{C}[1]{>{\centering\arraybackslash}p{#1}}

\usepackage[nolist]{acronym}        
\usepackage{verbatim}               
\usepackage{comment}                

\title{Improving Reproducible Deep Learning Workflows with DeepDIVA}

\author{
    \IEEEauthorblockN{
        \textbf{Michele~Alberti}\IEEEauthorrefmark{1}\IEEEauthorrefmark{2}, \and
        \textbf{Vinaychandran~Pondenkandath}\IEEEauthorrefmark{1}\IEEEauthorrefmark{2}, \and
        \thanks{\IEEEauthorrefmark{1} Both authors contributed equally to this work.}
        \textbf{Lars~V{\"o}gtlin}\IEEEauthorrefmark{2}, \and
        \textbf{Marcel~W{\"u}rsch}\IEEEauthorrefmark{2}\IEEEauthorrefmark{3}, \and
        \textbf{Rolf~Ingold}\IEEEauthorrefmark{2}, \and
        \textbf{Marcus~Liwicki}\IEEEauthorrefmark{2}\IEEEauthorrefmark{4}
    }\\
    \vspace{0.2cm}
    \IEEEauthorblockA{
        \IEEEauthorrefmark{2}%
        \textit{Document Image and Voice Analysis Group (DIVA)} \\
        University of Fribourg, Switzerland\\
        \{firstname\}.\{lastname\}@unifr.ch \\
        \vspace{0.15cm}
        \IEEEauthorrefmark{3}%
        \textit{Institute for Interactive Technologies (IIT)} \\
        FHNW University of Applied Sciences and Arts Northwestern Switzerland, Switzerland\\
        marcel.wuersch@fhnw.ch\\
        \vspace{0.15cm}
        \IEEEauthorrefmark{4}%
        \textit{EISLAB Machine Learning} \\
        Lulea University of Technology, Sweden\\
        marcus.liwicki@ltu.se\\
    }
}

\begin{document}

\maketitle

\thispagestyle{empty}

\begin{acronym}[Bash]
    \acro{DD}{DeepDIVA}
\end{acronym}


\begin{abstract}

The field of deep learning is experiencing a trend towards producing reproducible research. 
Nevertheless, it is still often a frustrating experience to reproduce scientific results.
This is especially true in the machine learning community, where it is considered acceptable to have black boxes in your experiments. 
We present \acl{DD}, a framework designed to facilitate easy experimentation and their reproduction. 
This framework allows researchers to share their experiments with others, while providing functionality that allows for easy experimentation, such as: boilerplate code, experiment management, hyper-parameter optimization, verification of data integrity and visualization of data and results. 
Additionally, the code of \acl{DD} is well-documented and supported by several tutorials that allow a new user to quickly familiarize themselves with the framework. 



\end{abstract}

\section{Introduction}
\label{toc:introduction}


In the context of science, reproducibility refers to the ability to reproduce the scientific results of other researchers. 
Science is a highly collaborative process, with most research built on the work of others.
Generating reproducible research is important, as it helps others to verify findings and build further experiments using shared code as starting points. 
In the recent years, experimental reproducibility -- and it's lack of -- has been a point of concern in the scientific community.
This problem -- also known as the reproducibility crisis or the replication crisis -- has been extensively studied in the fields of psychology~\cite{john2012measuring}, economics~\cite{camerer2016evaluating} and medicine~\cite{begley2013reproducibility}. 
Recently, \cite{hutson2018artificial} and \cite{olorisade2017reproducibility} have shown that reproducibility is a problem in the field of machine learning as well. 

Machine learning research typically involves several rounds of experimentation, which have challenges such as inherent randomness due to seeds, large number of hyper-parameters, incremental changes to the code and often experiments are run in various environments with variations in software and hardware between experiments.
On occasion, details that are necessary to reproduce experiments are not included due to confidentiality requirements and space constraints in the papers. 
In addition, due to the pressure of the publish-or-perish nature of academia, researchers are incentivized to spend their time trying new ideas or different experiments instead of building robust infrastructure to support rigorous experimentation. 
Therefore, the code that is written for such experiments is of poor quality, and most researchers are reluctant to share such code. 
In the cases when such code is shared, the poor quality of the code makes it difficult to read or use and therefore reproduce. 

\begin{table}[t]
  \caption{Analysis of 50 Reinforcment Learning publications at major Machine Learning conferences. These results allow to draw the conclusion that reproducibility is still difficult, through a combination of source codes not available, not reporting all hyper-parameters, and not reporting the hyper-parameter selection method. (Source: video recording of the talk\textsuperscript{\ref{note1}})}
\label{table:reproducibility_nips}
\begin{center}
\begin{small}
\begin{sc}
\resizebox{0.48\textwidth}{!}{%
\begin{tabular}{@{}ll@{}}
\toprule
                                                     & Yes   \\ \midrule
Paper has experiments                                & 100\% \\
Paper uses neural networks                           & 90\%  \\
All hyperparams for proposed algorithm are provided  & 90\%  \\
All hyperparams for baselines are provided           & 60\%  \\
Code is linked                                       & 55\%  \\
Method for choosing hyperparams is specified         & 20\%  \\
Evaluations on some variation of a hold-out test set & 10\%  \\
Significance testing applied                         & 5\%   \\ \bottomrule
\end{tabular}%
}
\end{sc}
\end{small}
\end{center}
\end{table}
 
\begin{table*}[!t]
\caption{
Examples of tasks for which the entire pipeline is already implemented in \acl{DD}. 
These tasks are inherently very different, especially their source domain e.g. images, videos, text and bi-dimensional data. 
Yet, their work-flow and the code infrastructure necessary for handling them is mostly similar. 
DeepDIVA leverages these similarities by implementing a modular structure, which, combined with the extensive documentation and the multiple tutorials available makes extending or modifying an existing task an easy and swift procedure.}
\label{table:dl_out_of_the_box}
\begin{center}
\begin{small}
\begin{sc}
\begin{tabular}{ccc c ccc}
    \toprule
    Input & Task & Output &\quad\quad & Input & Task & Output \\ 
    \midrule
    
    \raisebox{-.5\height}{\includegraphics[height=1.7cm]{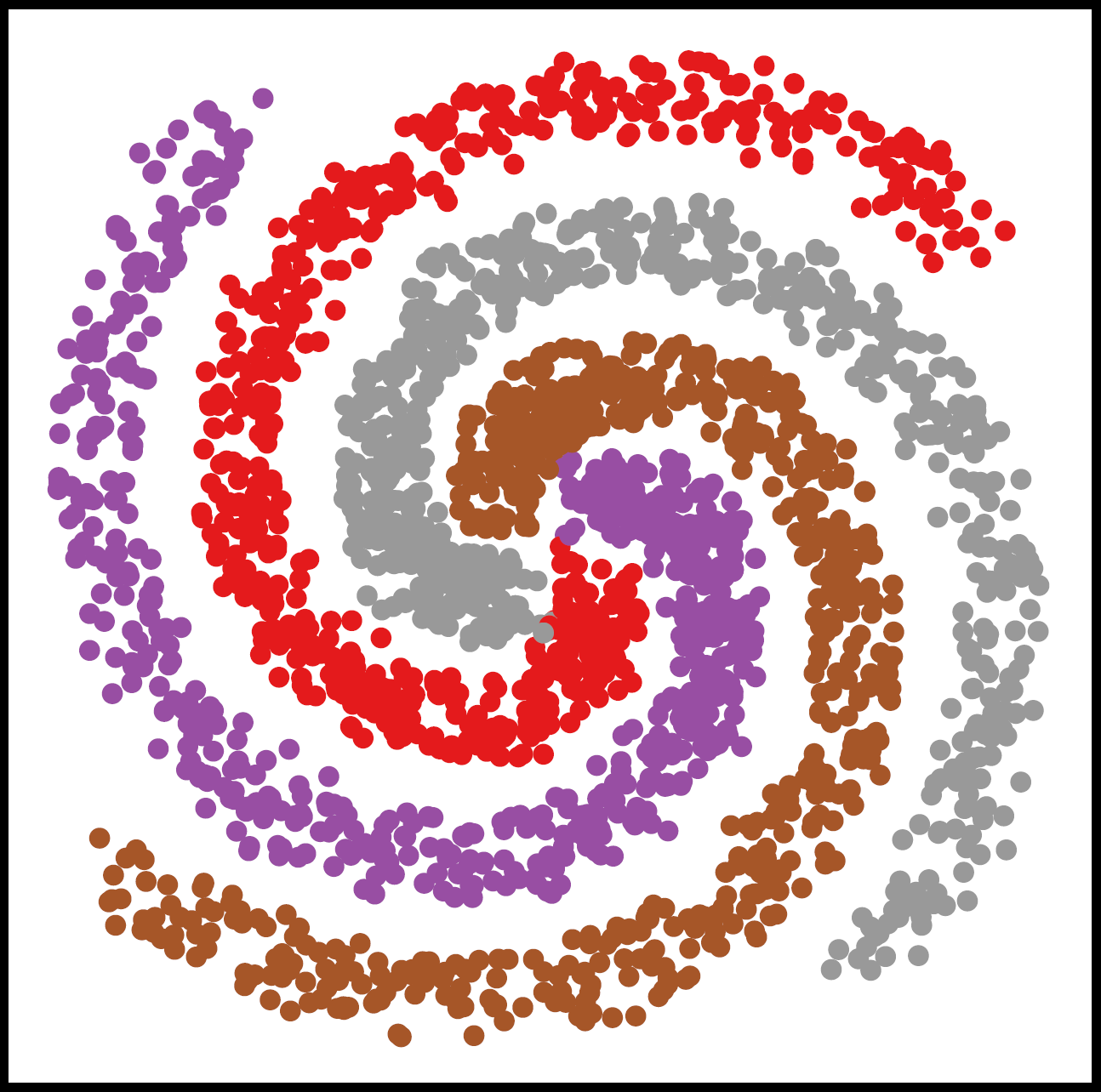}}
    & 
    \shortstack{Two \\ Dimensional}     
    & 
    \raisebox{-.5\height}{\includegraphics[height=1.7cm]{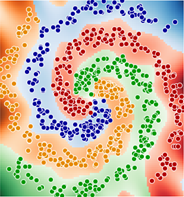}}
    &&
    \raisebox{-.5\height}{\includegraphics[height=1.7cm]{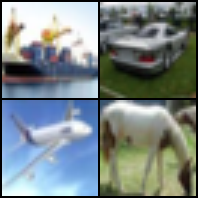}}
    & 
    \shortstack{Image \\ Auto-Encoding}     
    & 
    \raisebox{-.5\height}{\includegraphics[height=1.7cm]{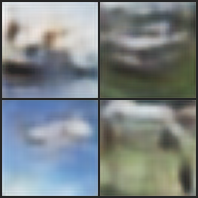}}
    \\ \\ 
    
    \raisebox{-.5\height}{\includegraphics[height=1.7cm]{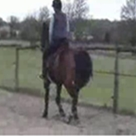}}
    & 
    \shortstack{Video Action \\ Recognition}     
    & 
    \raisebox{-.5\height}{\includegraphics[height=1.7cm]{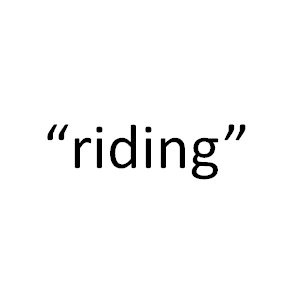}}
    &&
    \raisebox{-.5\height}{\includegraphics[height=1.7cm]{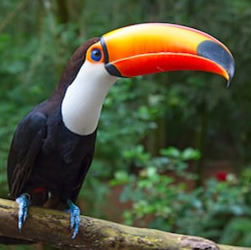}}
    & 
    \shortstack{Image \\ Similarity}    
    & 
    \raisebox{-.5\height}{\includegraphics[height=1.7cm]{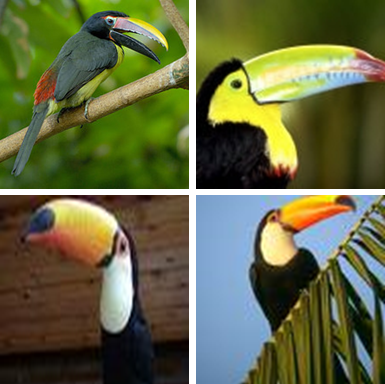}}
    \\ \\
    
    \raisebox{-.5\height}{\includegraphics[height=1.7cm]{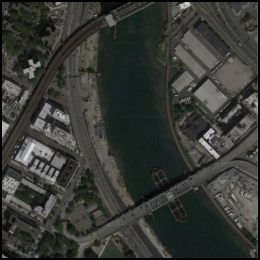}}
    & 
    \shortstack{Semantic \\ Segmentation}    
    & 
    \raisebox{-.5\height}{\includegraphics[height=1.7cm]{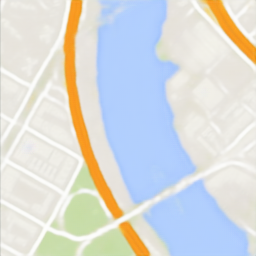}}
    &&
    \raisebox{-.5\height}{\includegraphics[height=1.7cm]{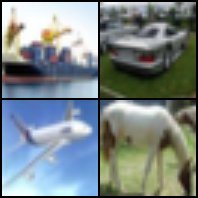}}
    & 
    \shortstack{Image \\ Classification}    
    & 
    \raisebox{-.5\height}{\includegraphics[height=1.7cm]{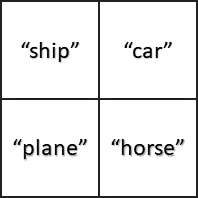}}
    \\ \\ 
    
    \raisebox{-.5\height}{\includegraphics[height=1.7cm]{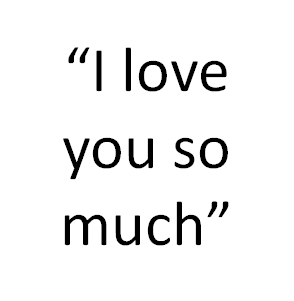}}
    & 
    \shortstack{Sentence \\ Classification}     
    & 
    \raisebox{-.5\height}{\includegraphics[height=1.7cm]{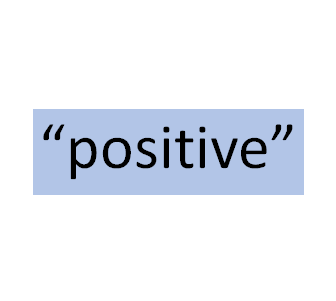}}
    &&
    \raisebox{-.5\height}{\includegraphics[height=1.7cm]{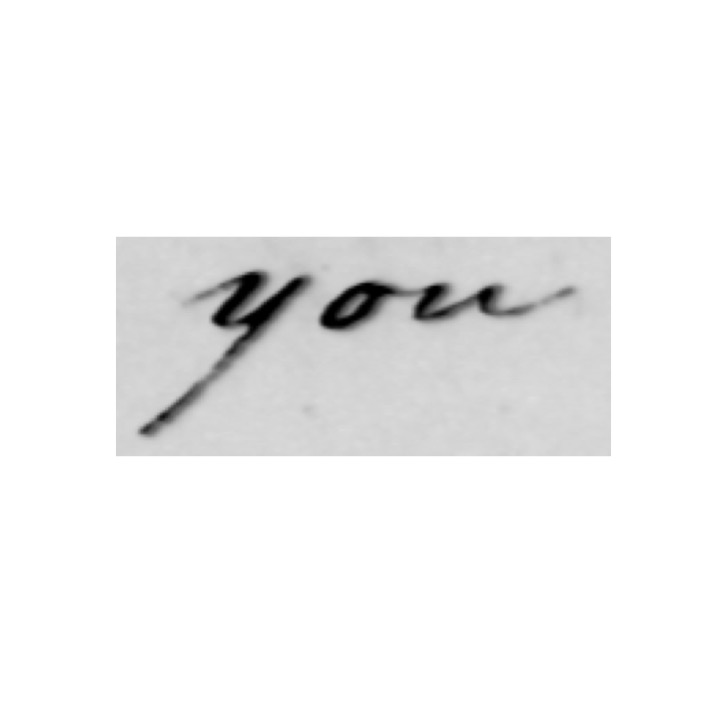}}
    & 
    \shortstack{Word \\ Spotting}     
    & 
    \raisebox{-.5\height}{\includegraphics[height=1.7cm]{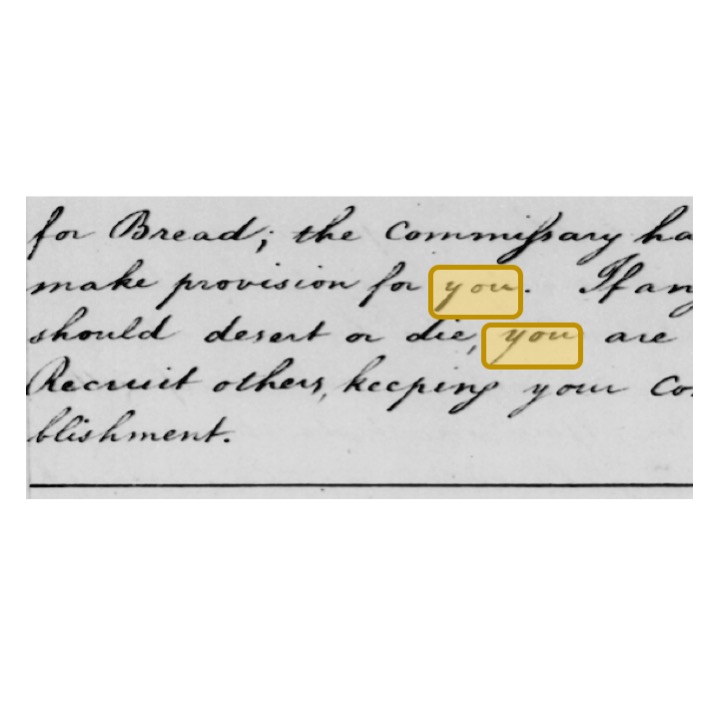}}
    \\
    
    \bottomrule
\end{tabular}
\end{sc}
\end{small}
\end{center}
\end{table*}

\subsection*{Reproducibility Crisis}
\label{toc:related work}
Reproducible Research has been a topic of discussion within research for several years now.
In 2015, Nature asked 1500 scientists~\cite{Baker2016a} across different domains to gather their opinions on reproducibility within their respective fields.
A large majority of 90\% answered that there is a significant (52\%), or a slight (38\%) crisis regarding this topic.
Unfortunately this study does not directly account for Computer Science or Machine Learning.

At NeurIPS 2018, Joelle Pineau had an invited talk on the topic of "Reproducible, Reusable, and Robust Reinforcement Learning"\footnote{\label{note1}see: \url{http://bit.ly/neurips_reproducibility}}.
In this talk she presented a table as shown in Table \ref{table:reproducibility_nips}.
In a study they analyzed 50 Reinforcment Learning papers presented at large Machine Learning conferences (NeurIPS, ICML, ICLR), showing that only a small majority of them provide source code to their publications.
Combined with the facts that neither all hyper-parameters are provided, nor the method for choosing them allows for the conclusions that reproducing the results of these papers would be a very difficult task

The abovementioned examples show that there is a need for tools that support researchers in performing reproducible research.
Such a tool should provide support for the following when running experiments:
\begin{itemize}
    \item Fixing and storing random seeds
    \item Ensuring that an exact version of the code is available
    \item Provide boilerplate code for performing experiments
\end{itemize}
Different tools have been proposed previously and we provide an overview of them in~\cite{albertipondenkandath2018deepdiva}.
These solution, however, often target only one or few aspects of a machine learning pipeline, whereas we aim at providing support for the entire work-flow. 

Moreover, \acl{DD} provides basic implementations for common tasks, supports various visualizations and addresses the situation where an algorithm has non-deterministic behavior e.g. because of random initialization.
Finally, unlike others attempts which aim to be platform or language independent, \acl{DD} relies on a working Python environment and specific settings, allowing it to be lightweight and enable using GPU hardware in a straightforward way. 

\subsection*{Main Contribution}
\label{toc:Main Contribution}

We aim to contribute towards the growing needs for reproducibility and openness in the machine learning community by providing \acl{DD}, a framework that tries to close the gap between good engineering practices and fast moving research work-flows.
While an early version of \acl{DD} built for the purposes of handwriting recognition has presented in~\cite{albertipondenkandath2018deepdiva}, this paper presents a much more extensive framework with a wide range of functionality and templates in machine learning tasks.

\acl{DD} allows for quick experimentation for a variety of common scenarios, such as image, video and language classification, similarity matching, image segmentation, image auto-encoding among other things. 
The framework has been built in a modular and easily extensible manner such that additional tasks and capabilities can be added without extensive efforts. 
\acl{DD} integrates popular tools such as TensorBoard\footnote{\url{https://www.tensorflow.org/programmers_guide/summaries_and_tensorboard}} (for aggregating all the visualizations and results produced) and SigOpt~\cite{sigopt} for hyper-parameter optimization. 
Finally, the documentation and tutorials provided help smoothen the learning curve for new users or contributors.

\section{Reproducing Experiments}
\label{toc:reproducing_experiments}

Often, reproducing the code of others is quite difficult because not all papers come with the sources necessary to make the experiments work. 
Even when the code is available, there are often issues with getting the code to run. 
We try to alleviate this problem by proposing to conduct all research inside the easy-to-configure \acl{DD} environment. 

\subsection{How It Is Done}
\label{toc:how_done}

The \acl{DD} environment can be set up using a one-click bash script (see Section~\ref{toc:Deep-learning out-of-the-box}).
With a functioning \acl{DD} environment, one can reproduce any experiment using: a link to the appropriate fork of \acl{DD}, the specific commit identifier of the particular experiment, and the list of parameters or that were used to run the experiment or a bash file that contains the exact commands to re-run the experiment. 

%

\subsubsection{Log Everything}

The framework saves logs that detail several facets of the training procedure, such as: experimental setup parameters, information about the training data, evaluation metrics, model parameters, and all visualizations generated during the training. 
In addition to all of this, the framework also makes a snapshot of the code-base (as seen at execution time) and stores it along with the logs. 
With all of this information, it's possible to analyze the training procedure after the process or use intermediate model representations for other purposes.
This can be quite helpful for experiments that take longer amounts of time to run. 

\subsubsection{Seed All Randomness}

Controlling for code and parameters is not enough to ensure reproducbility in machine learning.
Many machine learning methods are randomly instantiated, and the results of such experiments are highly subject to the instantiation. 
To make a perfect reproduction, or indeed to compare the effect of methods or parameters, it is necessary to be able to remove all sources of randomness from the experiment. 
DeepDIVA allows the user to specify a seed, upon which all sources of randomness in the system are controlled, allowing for perfect reproducibility. 

\subsubsection{Enforcing Version Control}
A scenario that most researchers are likely familiar with is the sudden inability to get results that you had a few code changes prior.
This can be due to changing hard-coded parameters, or (un)commenting lines of code to change execution flow.
We aim to tackle this scenario by enforcing the user to commit their code before running any experiments. 
The framework checks before running an experiment if the user has checked in and committed their code. 
However, this can be annoying and tedious during the development process to commit all small changes before running experiments.
In this case, the backup solution of the framework activates and makes a copy of all the source code in repository in the log files of the experiment. 
\subsubsection{Data Integrity Management}
\label{toc:data_integrity}

In order to ensure full reproducibility, one requires access to the same data.
Since the collection, storage and dissemination of datasets is beyond the scope of the framework, it becomes necessary to have a way to ensure that one is in possession of the exact same data as the experiment to be reproduced. 
This feature was highly requested by the community since the frameworks initial release.

In \acl{DD} we implemented the verification through the use of a footprint file. 
The creation of this footprint is automated and happens immediately at the start of an experiment (if the file has not been generated before).
The content of the file is a JSON tree which stores the entire dataset structure in great detail ,i.e., it stores all file names and their SHA-1 hashes.
Moreover, there is a global ``last modified'' tag which contains the most recent value for the entire folder (spanning every file contained in it recursively). 
This tag is used at run-time to verify if the dataset has been modified since the footprint generation.
This check as one can imagine is very quick and hence does not affect the regular flow or run-time of an experiment.
However, this type of verification, albeit quick, is not secure against a malicious manipulation of the data, since a skilled attacker might modify it in subtle ways \cite{alberti2018tampering} and tamper the time stamp on the file system too.
To combat this --- very remote --- threat, there is the possibility to activate a deep inspection of the dataset integrity using the stored SHA-1 hashes.
In this way we can ensure that if the dataset integrity verification is successful, one is sure that there are no differences between the data on the file system and the dataset described by the footprint.

\section{Productivity Out-of-the-Box}
\label{toc:Deep-learning out-of-the-box}

Most researchers have established workflows which they are are comfortable with, however, these workflows may not be best-practices compliant and it can be quite difficult to change the way you do things in order to achieve the ever-increasing best-practices ideals of the field. 
Therefore, we try to make the experience of using \acl{DD} for the first time a quick simple and painless one, and allow researchers to be operational and productive as soon as possible. 

\acl{DD} is very easy to setup on MacOS and Ubuntu. 
Once the repository has been cloned, setting up a fully functional environment is a single bash script away.
Once set up, the framework has boilerplate code for several different scenarios (as seen in Table~\ref{table:dl_out_of_the_box}), notably: word-spotting, similarity matching for image,  classification of images, natural language, video and bi-dimensional data.

These templates cover many common scenarios that researchers often encounter, and as the scenarios are written in a modular fashion, they can be easily adapted or extended to other tasks in a quick and painless manner. 
This helps a researcher be quickly productive as implementing the boilerplate constitutes a significant part of the development process. 
Starting from an already implemented task and adapting it for a specific purpose allows researchers to avoid developing redundant code. 
The following sections recap several features that support a typical research workflow in \acl{DD} (introduced in \cite{albertipondenkandath2018deepdiva}).

\begin{figure*}[!t]
  \centering
  \subfloat[%
  Example of comparison accuracy for two different training protocols (orange and pink) on a classification task. 
  The visualization tool is not limited to two instances and allows for comparing an arbitrary number of instances.
  AnonymousFramework automatically measures the batch-wise (not shown in Figure) and epoch-wise loss and accuracy and plots in Tensorboard.]%
  {\includegraphics[width=.47\textwidth]%
  {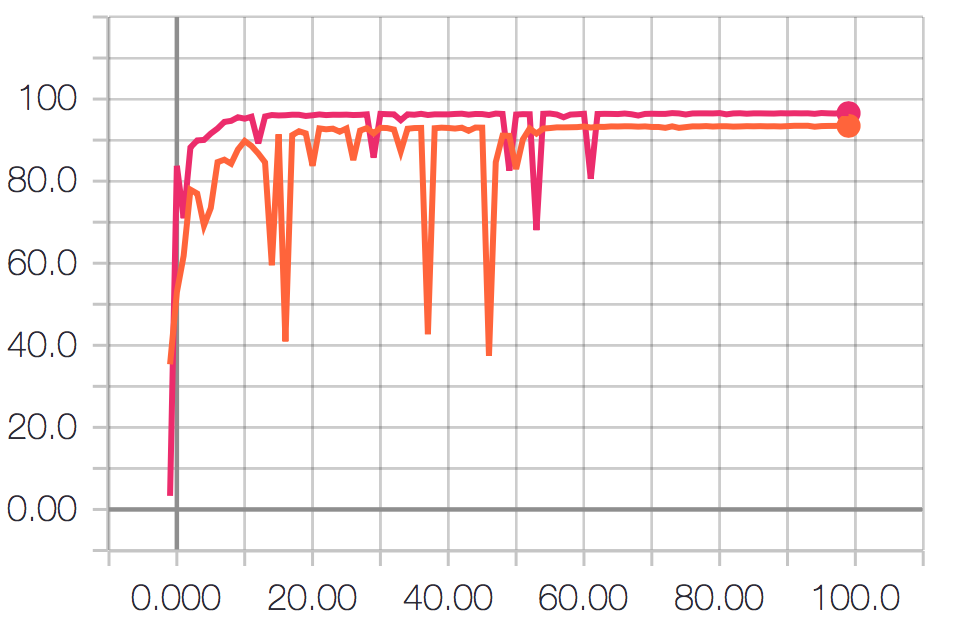}\label{subfig:compareRuns_watermarks}}
  \hfil
  \subfloat[%
  In all the experiments which involve randomness it is useful to evaluate how that affects the results obtained.
  Considering that very often in deep learning the networks are initialized with random weights, this is a common scenario.
  In this figure is shown an evaluation of how randomness affects execution by visualizing the aggregated results of multiple runs. 
  Here the full line represent the mean value, the dotted lines are the highest and the lowest results obtained. 
  Finally, the shaded area indicates the variance over all runs.]%
  {\includegraphics[width=.47\textwidth]%
  {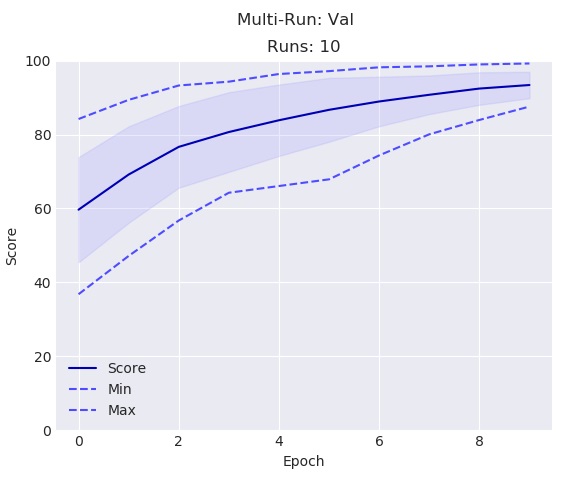}\label{subfig:shadyPlot}}
  \vfil
  \subfloat[%
  Confusion matrices are a well established tool for visualizing the performance of a system both in a binary and in a multi-class setting. %
  Our framework produces a confusion matrix every time the model is validated and finally when it is tested. 
  In the above figure is shown a confusion matrix for a 4 classes task, where the darker the color the higher the amount of samples was classified as such.
  Ideally the confusion matrix should look as full color on the diagonal and white everywhere else.]%
  {\includegraphics[width=.47\textwidth]%
  {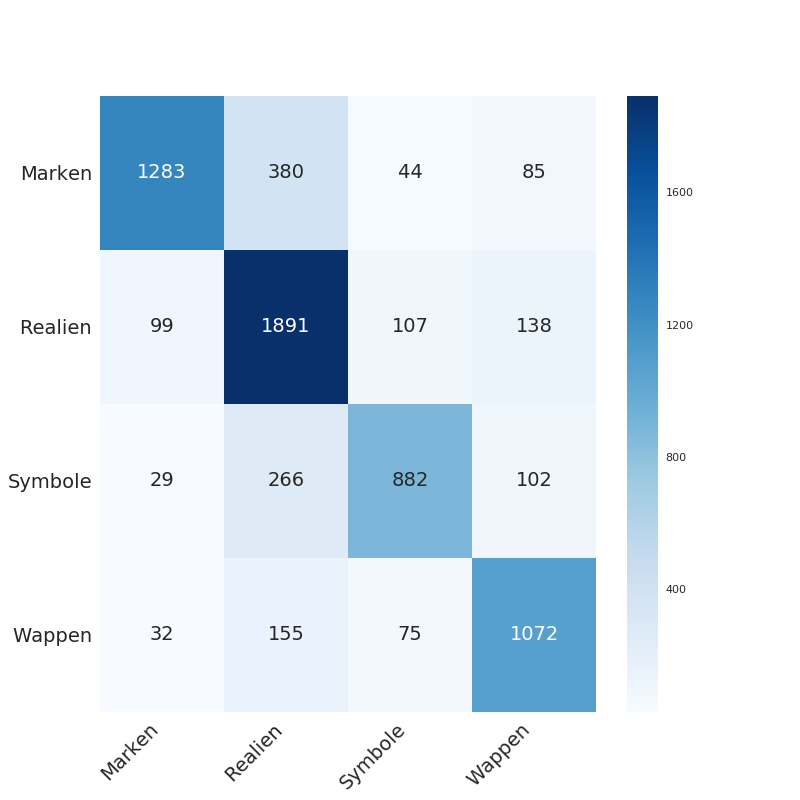}\label{subfig:confusionMatrix}}
  \hfil
  \subfloat[%
  Visualizing features is common step in a deep learning research pipeline as it often provides an insight on the model and/or on the data one is working with.
  We integrated the native feature visualization of Tensorboard into the framework. 
  Specifically one can choose to use the 
  T-Distributed Stochastic Neighbor Embedding (T-SNE)~\cite{maaten2008visualizing} or the Principal Component Analysis (PCA) to project the high-dimensional features embedding in a 2D or 3D space.
  In this figure is shown an example of T-SNE feature projection of CIFAR-10.]%
  {\includegraphics[width=.47\textwidth]%
  {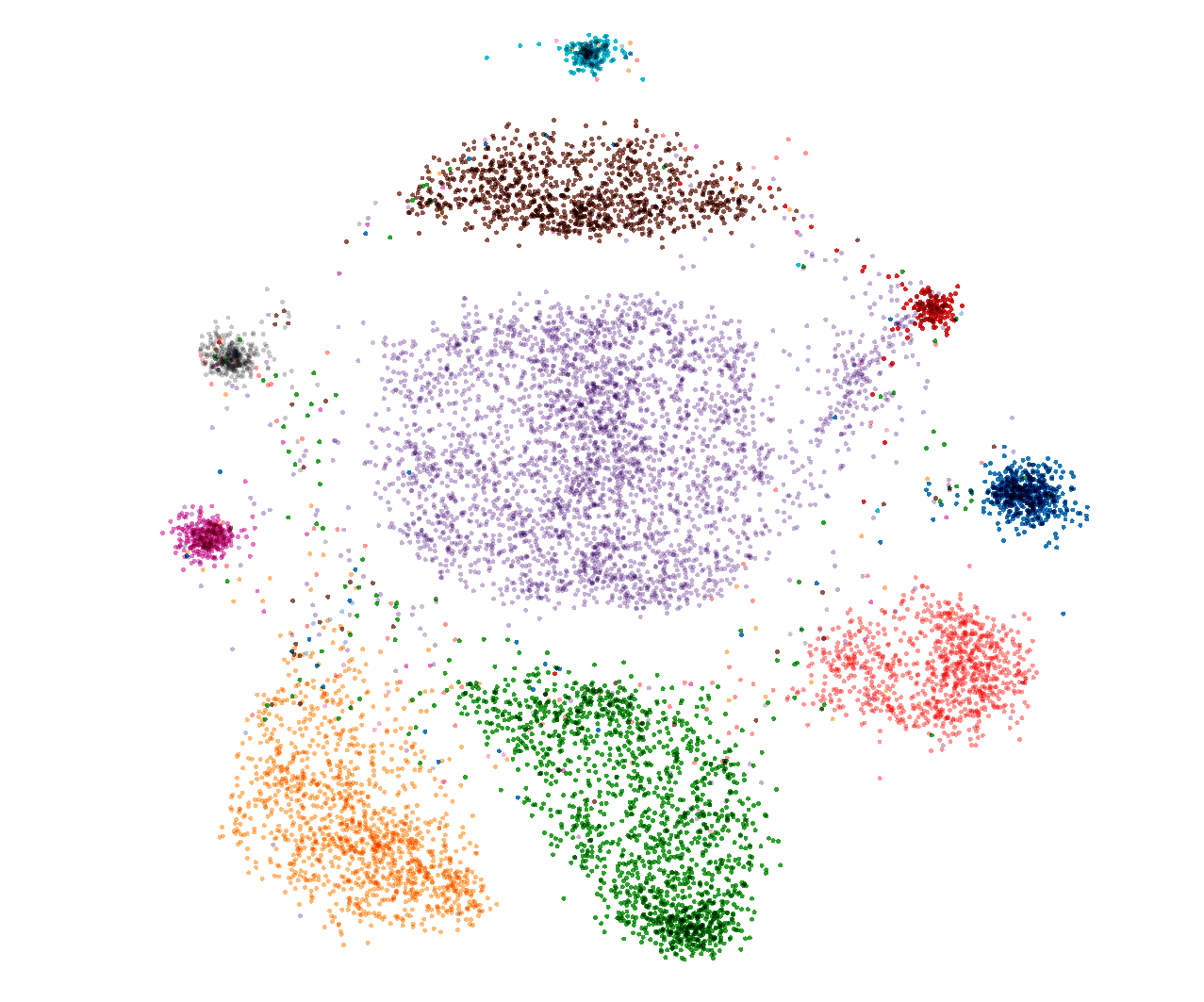}\label{subfig:tsne_watermarks}}
  \caption{%
  In this figure are shown some examples of different visualizations produced automatically by \acl{DD}.
  This is not an exhaustive list, but many other visualizations are task-specific and might required a significant amount of context to be understood.
  All visualization as available in real-time as the training progresses. 
  This is a point which we believe to be critically important, as it allows to take important decisions before the end of the experiments thus saving an conspicuous amount time. 
  Credit for the figures to~ \cite{albertipondenkandath2018deepdiva}.}
  \label{fig:visualization_produced2}
\end{figure*}






\subsection{Prepare Your Data} 

The first step in any machine learning task is acquiring and preparing the data. 
\acl{DD} comes equipped with some tools to support this task.

\begin{itemize}
    \item \textbf{Download a dataset with a click} \acl{DD} supports downloading and preparing several datasets (CIFAR-10~\cite{cifar}, MNIST~\cite{mnist}, SVHN~\cite{svhn}, STL-10~\cite{stl}) with a single command. 
    \item \textbf{Split your dataset} The framework contains a script to split an arbitrary dataset (stored in a standard format) into classic machine learning splits.
    \item \textbf{Analyze the data}  Computing the mean, standard deviation and class distribution for pre-processing the data is a standard operation. \acl{DD} provides scripts to compute this for large datasets in an online manner. 
    \item \textbf{Ensure data integrity} \acl{DD} keeps track of your datasets to ensure that they are not modified. More details in Section~\ref{toc:data_integrity}
\end{itemize}

Once your data has been downloaded or prepared in the correct format, \acl{DD} loads and pre-processes the datasets such that they can be used in the appropriate tasks.
    
\subsection{See What Your Network Thinks With 2D Data}
During the research process for a new idea, you might want to try out your idea on a simple toy-dataset before progressing to more complex datasets. 
To enable you to do so, \acl{DD} offers a workflow to test out your idea on bi-dimensional data which allows you can visualize exactly what your network thinks of the output space.
\acl{DD} contains all the necessary code to perform such a visual analysis of the network, and all a user needs to do is to modify the task and implement their research idea.  
\vspace{-3mm}
\subsection{Real-Time Data Visualization:}
We use the Tensorboard application developed by TensorFlow~\cite{tensorflow} to aggregate all the visualizations produced by the framework. 
Normal training and validation curves are plotted directly, and all other visualization produced by the framework are added directly into images section of the corresponding experiment. 
\acl{DD} dynamically generates plots for executions (see Fig.~\ref{subfig:compareRuns_watermarks}) and makes them available in Tensorboard, thus experiments with differing configurations can be compared, as well as performance of two or more methods. 
The multi-run flag automatically reruns an experiment a given number of times and aggregates the result into a plot (see Fig.~\ref{subfig:shadyPlot}). 
\acl{DD} also generates a confusion matrix during evaluation (see Fig.~\ref{subfig:confusionMatrix}). 


\subsection{Automatic Hyper-parameter Optimization:}
Instead of having to perform the tedious and time-consuming procedure of optimizing hyper-parameters by hand, the researcher can simply use a single command line parameter and let the framework deal with it thanks to SigOpt~\cite{sigopt} integration.


\section{Be a Part Of It}
\label{toc:join}


Many of the available tools are extremely good at what they are designed for, however they often have steep learning curves. 
Even during the setup phase, several tools expect a user to have the skill, time and patience to set up an environment manually. 
Indeed, the authors of this paper have even had the experience of encountering Github repositories where the setup instructions are to simply install packages as you encounter errors. 
This often discourages the average user and significantly increases the time required to get to a productive stage. 

Additionally, the quality of the documentation or tutorials (or lack thereof) determine the impact of a tool, no matter how effective it may be. 
When this is combined with stringent contribution guidelines, or lack of an open-source nature, it can render a tool community-unfriendly. 
This is a major issue for the field as the quality of a framework is measured not only by the quality of the results delivered by it, but also by it's maintenance, the learning curve and the adoption overhead. 



To foster a friendly and productive community of researchers, we try to make \acl{DD} accessible by tackling the aforementioned problems as follow:

\textbf{No Setup Time:} 
\acl{DD} can be setup with a single bash script on both Ubunutu and MacOS. (see Section~\ref{toc:how_done})

\textbf{Documentation:}
The framework is documented\footnote{See the documentation at  \url{link\_redacted\_for\_blind\_submission}} such that it can be used in a educational environment for didactic purposes.


\textbf{Tutorials:}
There is a friendly ``Getting started'' followed by a plethora of tutorials \footnote{See tutorials at  \url{link\_redacted\_for\_blind\_submission}} which will help a new user learn and use the available features efficiently. 
For example, there are tutorials on how to prepare the data, load it and run the implemented tasks (see \ref{table:dl_out_of_the_box}) as well as how to visualize the results.
More experienced people can also find tutorials on how to extend the framework and perform advanced operations with it. 
These tutorials are not intended to teach someone machine learning, but rather how to use \acl{DD} to do achieve their ideas better.

\textbf{Fork It\footnote{See the repository at \url{link\_redacted\_for\_blind\_submission}}:}
\acl{DD} is built with the goal of being extensible and modular. 
It is open-source and comes with verbose documentation such that the core code will be accessible to everyone.
It has been designed in a modular way which favors and encourages growth and modifications, in contrast with other solutions which optimize performance at the expense of maintenance 
Moreover, being a collaborative project, additions suggested by users can be integrated benefiting the community as a whole.
This is not always possible or can be difficult due to closed source software, commercial solutions or impenetrable core code.

\section{Conclusion and Future Work}

We contribute towards meeting the demands for reproducibility and openness in machine learning by providing \acl{DD}: an open-source Python deep-learning framework designed to enable quick and intuitive setup of reproducible experiments with a large range of useful analysis functionality.
We show how researchers can quickly include it in their workflow (thanks to detailed documentation and easy tutorials) thus saving time while enabling reproducing their research in a quick and intuitive fashion.
In the near future \acl{DD} will include more visualization tools, provided by the small (but thriving) community of developers which is forming around it.

\section*{Acknowledgment}
The work presented in this paper has been partially supported by the HisDoc III project funded by the Swiss National Science Foundation with the grant number $205120$\textunderscore$169618$.

\bibliographystyle{IEEEtran}
\bibliography{biblio}

\begin{thebibliography}{10}
\providecommand{\url}[1]{#1}
\csname url@samestyle\endcsname
\providecommand{\newblock}{\relax}
\providecommand{\bibinfo}[2]{#2}
\providecommand{\BIBentrySTDinterwordspacing}{\spaceskip=0pt\relax}
\providecommand{\BIBentryALTinterwordstretchfactor}{4}
\providecommand{\BIBentryALTinterwordspacing}{\spaceskip=\fontdimen2\font plus
\BIBentryALTinterwordstretchfactor\fontdimen3\font minus
  \fontdimen4\font\relax}
\providecommand{\BIBforeignlanguage}[2]{{%
\expandafter\ifx\csname l@#1\endcsname\relax
\typeout{** WARNING: IEEEtran.bst: No hyphenation pattern has been}%
\typeout{** loaded for the language `#1'. Using the pattern for}%
\typeout{** the default language instead.}%
\else
\language=\csname l@#1\endcsname
\fi
#2}}
\providecommand{\BIBdecl}{\relax}
\BIBdecl

\bibitem{john2012measuring}
L.~K. John, G.~Loewenstein, and D.~Prelec, ``Measuring the prevalence of
  questionable research practices with incentives for truth telling,''
  \emph{Psychological science}, vol.~23, no.~5, pp. 524--532, 2012.

\bibitem{camerer2016evaluating}
C.~F. Camerer, A.~Dreber, E.~Forsell, T.-H. Ho, J.~Huber, M.~Johannesson,
  M.~Kirchler, J.~Almenberg, A.~Altmejd, T.~Chan \emph{et~al.}, ``Evaluating
  replicability of laboratory experiments in economics,'' \emph{Science}, vol.
  351, no. 6280, pp. 1433--1436, 2016.

\bibitem{begley2013reproducibility}
C.~G. Begley, ``Reproducibility: six red flags for suspect work,''
  \emph{Nature}, vol. 497, no. 7450, p. 433, 2013.

\bibitem{hutson2018artificial}
M.~Hutson, ``Artificial intelligence faces reproducibility crisis,''
  \emph{American Association for the Advancement of Science}, 2018.

\bibitem{olorisade2017reproducibility}
B.~K. Olorisade, P.~Brereton, and P.~Andras, ``Reproducibility in machine
  learning-based studies: An example of text mining,'' 2017.

\bibitem{Baker2016a}
\BIBentryALTinterwordspacing
M.~Baker, ``1,500 scientists lift the lid on reproducibility,'' \emph{Nature},
  vol. 533, no. 7604, pp. 452--454, may 2016. [Online]. Available:
  \url{http://www.nature.com/doifinder/10.1038/533452a}
\BIBentrySTDinterwordspacing

\bibitem{albertipondenkandath2018deepdiva}
Anonymous, ``{AnonymousFramework},'' 2018.

\bibitem{sigopt}
\BIBentryALTinterwordspacing
I.~SigOpt, ``Sigopt reference manual,'' 2014. [Online]. Available:
  \url{http://www.sigopt.com}
\BIBentrySTDinterwordspacing

\bibitem{alberti2018tampering}
M.~Alberti, V.~Pondenkandath, M.~W\"ursch, M.~Bouillon, M.~Seuret, R.~Ingold,
  and M.~Liwicki, ``{Are You Tampering With My Data?}'' in \emph{2018 15th
  European Conference on Computer Vision (ECCV), Workshop on Objectionable
  Content and Misinformation (WOCM)}, Munich, Germany, sep 2018.

\bibitem{maaten2008visualizing}
L.~v.~d. Maaten and G.~Hinton, ``Visualizing data using t-sne,'' \emph{Journal
  of machine learning research}, vol.~9, no. Nov, pp. 2579--2605, 2008.

\bibitem{cifar}
A.~Krizhevsky, V.~Nair, and G.~Hinton, ``The {CIFAR}-10 dataset,''
  \emph{online: http://www. cs. toronto. edu/kriz/cifar. html}, 2014.

\bibitem{mnist}
Y.~LeCun, ``The mnist database of handwritten digits,'' \emph{http://yann.
  lecun. com/exdb/mnist/}, 1998.

\bibitem{svhn}
Y.~Netzer, T.~Wang, A.~Coates, A.~Bissacco, B.~Wu, and A.~Y. Ng, ``Reading
  digits in natural images with unsupervised feature learning,'' in \emph{NIPS
  workshop on deep learning and unsupervised feature learning}, vol. 2011,
  no.~2, 2011, p.~5.

\bibitem{stl}
A.~Coates, A.~Ng, and H.~Lee, ``An analysis of single-layer networks in
  unsupervised feature learning,'' in \emph{Proceedings of the fourteenth
  international conference on artificial intelligence and statistics}, 2011,
  pp. 215--223.

\bibitem{tensorflow}
\BIBentryALTinterwordspacing
M.~Abadi, P.~Barham, J.~Chen, Z.~Chen, A.~Davis, J.~Dean, M.~Devin,
  S.~Ghemawat, G.~Irving, M.~Isard, M.~Kudlur, J.~Levenberg, R.~Monga,
  S.~Moore, D.~G. Murray, B.~Steiner, P.~Tucker, V.~Vasudevan, P.~Warden,
  M.~Wicke, Y.~Yu, and X.~Zheng, ``Tensorflow: A system for large-scale machine
  learning,'' in \emph{12th USENIX Symposium on Operating Systems Design and
  Implementation (OSDI 16)}, 2016, pp. 265--283. [Online]. Available:
  \url{https://www.usenix.org/system/files/conference/osdi16/osdi16-abadi.pdf}
\BIBentrySTDinterwordspacing

\end{thebibliography}

\end{document}